\documentclass[journal]{IEEEtran}

\usepackage{cite}
\ifCLASSINFOpdf
  \usepackage[pdftex]{graphicx}
  \graphicspath{{../pdf/}{../jpeg/}}
  \DeclareGraphicsExtensions{.pdf,.jpeg,.png}
\else
  \usepackage[dvips]{graphicx}
  \graphicspath{{../eps/}}
  \DeclareGraphicsExtensions{.eps}
\fi
\usepackage{amsmath}
\usepackage{amssymb}
\usepackage{amsfonts}
\usepackage{multirow}
\usepackage{booktabs}
\usepackage{tabularx}
\usepackage{subfloat}
\usepackage{epstopdf}
\usepackage[labelfont=bf,textfont=bf]{caption}
\usepackage[caption=false,font=normalsize,labelfont=sf,textfont=sf]{subfig}

\usepackage{algorithm}  
\usepackage{algpseudocode}
\usepackage{epsfig}
\usepackage{bbm}
\usepackage{makecell}
\usepackage{bm}
\usepackage{pifont}
\usepackage{bbding}
\usepackage{color}
\usepackage{xcolor}

\usepackage{array}
\usepackage{fixltx2e}
\usepackage{stfloats}
\usepackage{url}
\begin{document}
%
% paper title
% Titles are generally capitalized except for words such as a, an, and, as,
% at, but, by, for, in, nor, of, on, or, the, to and up, which are usually
% not capitalized unless they are the first or last word of the title.
% Linebreaks \\ can be used within to get better formatting as desired.
% Do not put math or special symbols in the title.
%\title{Exploiting Spatio-Temporal Relation via Local-Global Transformer for Text-Video Retrieval}

\title{Efficient Cross-Modal Video Retrieval with Meta-Optimized Frames}

%
%
% author names and IEEE memberships
% note positions of commas and nonbreaking spaces ( ~ ) LaTeX will not break
% a structure at a ~ so this keeps an author's name from being broken across
% two lines.
% use \thanks{} to gain access to the first footnote area
% a separate \thanks must be used for each paragraph as LaTeX2e's \thanks
% was not built to handle multiple paragraphs
%

\author{Ning~Han, Xun Yang, Ee-Peng Lim, Hao Chen, Qianru Sun % <-this % stops a space

\thanks{Ning Han and Hao Chen are with the Department of Information Science and Engineering, Hunan University, Changsha 410082, China. E-mail: \{ninghan, chenhao \}@hnu.edu.cn.}% <-this % stops a space
\thanks{Xun Yang is with the School of Information Science and Technology, University of Science and Technology of China, Hefei 230026, China. E-mail:xyang21@ustc.edu.cn.}
\thanks{Ee-Peng Lim and Qianru Sun are with the School of Computing and Information Systems, Singapore Management University, Singapore 178902. E-mail: \{eplim, qianrusun \}@smu.edu.sg.}

}
 % <-this % stops a space

%%IEEE TRANSACTIONS ON MULTIMEDIA
% The paper headers
\markboth{}%
{Shell \MakeLowercase{\textit{et al.}}: Bare Demo of IEEEtran.cls for IEEE Journals}
% The only time the second header will appear is for the odd numbered pages
% after the title page when using the twoside option.
% 
% *** Note that you probably will NOT want to include the author's ***
% *** name in the headers of peer review papers.                   ***
% You can use \ifCLASSOPTIONpeerreview for conditional compilation here if
% you desire.

% If you want to put a publisher's ID mark on the page you can do it like
% this:
%\IEEEpubid{0000--0000/00\$00.00~\copyright~2015 IEEE}
% Remember, if you use this you must call \IEEEpubidadjcol in the second
% column for its text to clear the IEEEpubid mark.

% use for special paper notices
%\IEEEspecialpapernotice{(Invited Paper)}

% make the title area
\maketitle

% As a general rule, do not put math, special symbols or citations
% in the abstract or keywords.

\begin{abstract}
Cross-modal video retrieval aims to retrieve the semantically relevant videos given a text as a query, and is one of the fundamental tasks in Multimedia. Most of top-performing methods primarily leverage Visual Transformer (ViT) to extract video features~\cite{lei2021less, bain2021frozen, wang2022object}, suffering from high computational complexity of ViT especially for encoding long videos. A common and simple solution is to uniformly sample a small number (say, 4 or 8) of frames from the video (instead of using the whole video) as input to ViT. The number of frames has a strong influence on the performance of ViT, e.g., using 8 frames performs better than using 4 frames yet needs more computational resources, resulting in a trade-off. To get free from this trade-off, this paper introduces an automatic video compression method based on a bilevel optimization program (BOP) consisting of both model-level (i.e., base-level) and frame-level (i.e., meta-level) optimizations. 
The model-level learns a cross-modal video retrieval model whose input is the ``compressed frames'' learned by frame-level optimization. In turn, the frame-level optimization is through gradient descent using the meta loss of video retrieval model computed on the whole video. We call this BOP method as well as the ``compressed frames'' as Meta-Optimized Frames (MOF). By incorporating MOF, the video retrieval model is able to utilize the information of whole videos (for training) while taking only a small number of input frames in actual implementation. The convergence of MOF is guaranteed by meta gradient descent algorithms. For evaluation, we conduct extensive experiments of cross-modal video retrieval on three large-scale benchmarks: MSR-VTT, MSVD, and DiDeMo. Our results show that MOF is a generic and efficient method to boost multiple baseline methods, and can achieve a new state-of-the-art performance. Our code is public at: \emph{\url{https://github.com/sch971111/MOF}}.

\end{abstract}

% Note that keywords are not normally used for peerreview papers.

% For peer review papers, you can put extra information on the cover
% page as needed:
% \ifCLASSOPTIONpeerreview
% \begin{center} \bfseries EDICS Category: 3-BBND \end{center}
% \fi
%
% For peerreview papers, this IEEEtran command inserts a page break and
% creates the second title. It will be ignored for other modes.
\IEEEpeerreviewmaketitle

\section{Introduction}

\begin{figure}[t]
   \center
   \includegraphics[width=0.47\textwidth]{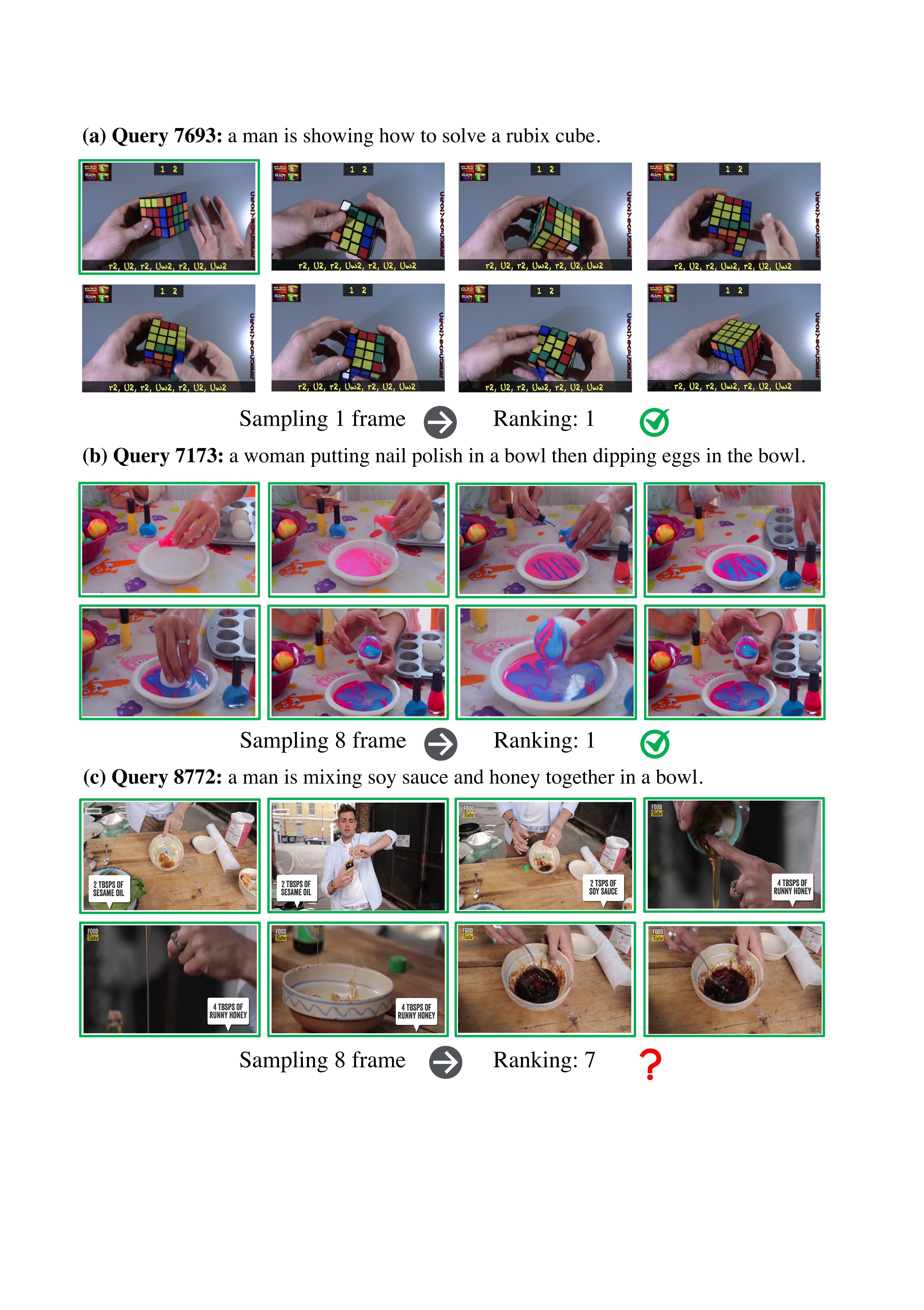}
   %\vspace{-0.1cm}
   \caption{Examples of easy (a), moderate (b) and hard (c) text-video pairs on the MSR-VTT dataset~\cite{xu2016msr}. 
   The green bounding box denotes sampled frames used for training a video retrieval model, e.g., Frozen~\cite{bain2021frozen}.
   The ranking of positive video is output by the trained model.
   %
%   The sampled positive frames are bounded by green boxes. The rank of the positive videos is returned by the trained retrieval model Frozen \cite{bain2021frozen} based on specific queries and varying numbers of frames. 
   %
%   We have two main observations. (1) Computational resources are wasted on applying the unnecessary 8 frames to ``easy'' video for which 1 frame are sufficient. (2) The ``hard'' text-video example is expected to sample more spatio-temporal informative frames useful for the corresponding text.}
    It is intuitive that easier (harder) samples need a smaller (larger) number of input frames. More frames may be needed for extremely hard samples, e.g., that in (c).
    % \qianru{please revise fig according to caption}
   }
   \label{fig:example}
   \vspace{-0.4cm}
\end{figure} 

\label{sec_intro}
\IEEEPARstart{T}{he} exponential growth of video data on the Internet has drawn a lot of attentions on video recognition \cite{feichtenhofer2019slowfast,lu2021learning,zhang2020hybrid,zhang2021token} and video retrieval tasks~\cite{chen2020interclass,li2020sea,qi2021semantics,wang2022siamese}.
The task of cross-modal video retrieval takes a text as a query and requires the model to search the most relevant videos based on this query. 
Its model usually contains two neural network modules---one for text encoding and one for video encoding~\cite{bain2021frozen,wang2022object}. The video encoding module is expected to learn spatial-temporal patterns in the video that can be align with the semantics in the query text. 
The state-of-the-art video encoder is based on Visual Transformer (ViT)~\cite{dosovitskiy2020image,bertasius2021space,liu2021video}, and is trained with a text encoding module in an end-to-end manner~\cite{devlin2019bert,sanh2019distilbert}. 
Nevertheless, ViT is costly when encoding long videos. Its complexity grows quadratically along with the number of input frames~\cite{bertasius2021space}. For example, in a video with $N$ frames, applying ViT on flattened spatio-temporal token sequences, where the length of the sequence is $S$ for each frame, will introduce an exponential complexity $O(N^2S^2)$ for both training and inference.
A simple and common solution is to uniformly sample a handful number of frames as input to ViT, instead of using the whole video. However, the ``sampling rate'' is a hyperparameter that has to be carefully selected. Besides, using sparse sampling causes a trade-off between effectiveness (i.e., retrieval accuracy) and efficiency (i.e., computing speed) for using ViT-based video retrieval models.

\begin{figure}[t]
   \center
   \includegraphics[width=0.49\textwidth]{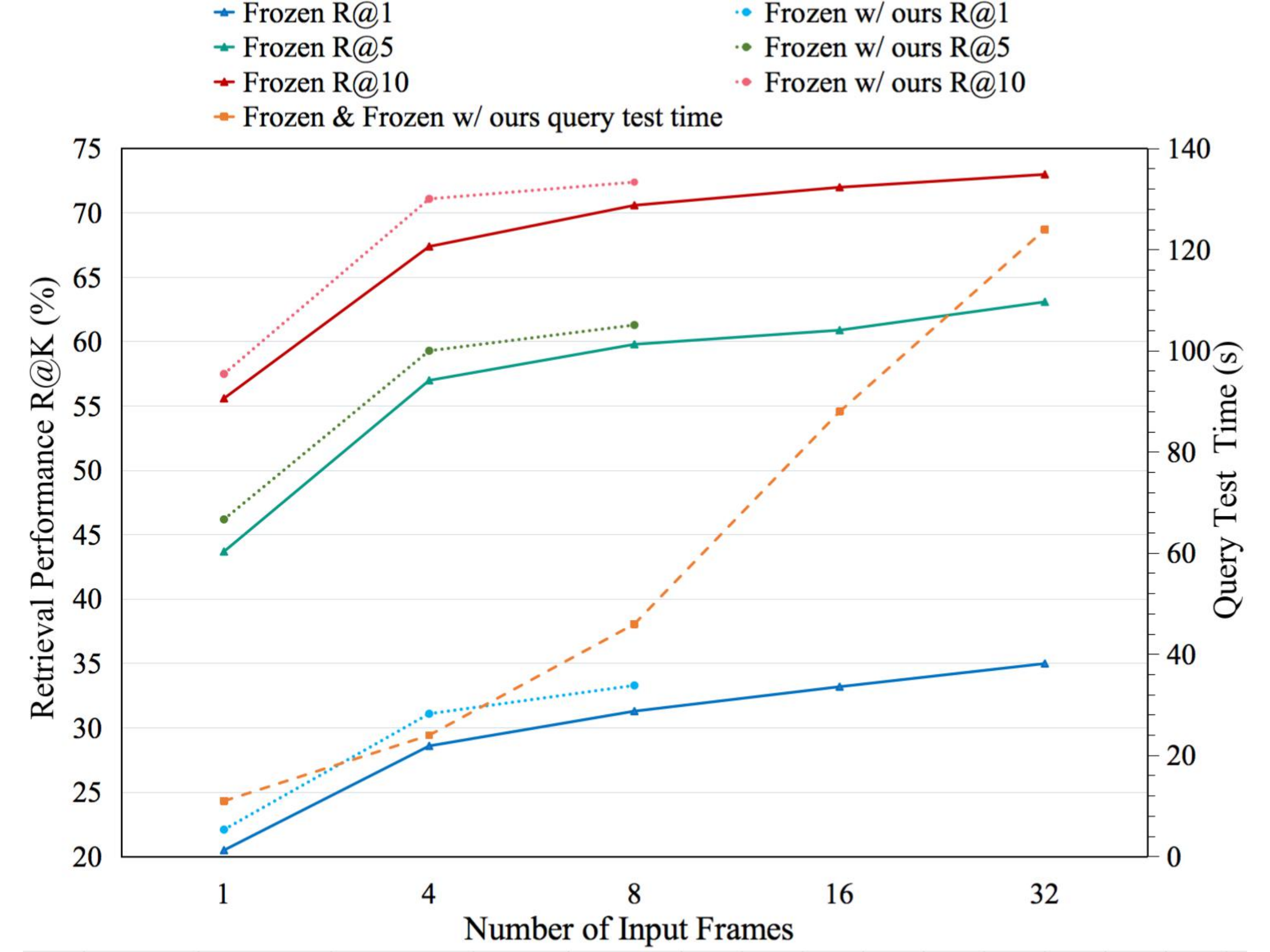}
   %\vspace{-0.3cm}
   \caption{Text-to-video retrieval performance (R@1, R@5, and R@10) and query test time on the MSR-VTT 1k-A test set \cite{Yu_2018_ECCV}. We train Frozen \cite{bain2021frozen} with varying frame numbers and Frozen \emph{w/} ours (i.e., using our MOF training as a plug-in module) with 16 frames compressed into different numbers of frames (please note that other compression rates are shown in the experiment section). We have two observations: (1) training with a large number of frames achieves better performance but a longer test time; and (2) Frozen \emph{w/} ours reaches better performances on different frame settings.}
%   tradeoff between accuracy and speed.}
   \label{fig:example_exp}
   %\vspace{-0.4cm}
\end{figure} 

We consider a few examples showing conventional methods, which sample frames as input of ViT. Bain et al. \cite{bain2021frozen} used a curriculum learning approach to sparsely sample frames in different training iterations. They limit the number of frames sampled per video to be only 4 or 8 and these sampled frames are passed to ViT in each iteration. 
However, this sampling method with a fixed frame number is sub-optimal and can not be adaptive to different-difficulty videos.
For example, in Figure~\ref{fig:example}, 
we show three video (and query) examples from easy to hard.
For the example in Figure~\ref{fig:example}(a), a single input frame from the video is enough to make a right prediction (for the model).
For the example in Figure~\ref{fig:example}(b), we need 8 input frames sampled from the video.
In contrast, for the example in Figure \ref{fig:example}(c), we can see that 8 input frames are still not enough for the model, i.e., the positive video is ranked 7th. 

To get a quantitative sense of the effect of frame number on model performance, we train the model using Frozen~\cite{bain2021frozen} (which is a top-performing method and is taken as our baseline) with varying frame numbers, and report the text-video retrieval performance and test time in Figure~\ref{fig:example_exp}. 
%
% The figure shows 
In particular, Figure~\ref{fig:example_exp} shows that training with a larger number of frames achieves better model performance but needs a significantly longer test time. Existing methods \cite{fan2021multiscale,wang2021efficient} for reducing this time cost are mostly focused on dropping redundant tokens which, however, brings a limited efficiency.

In contrast, our idea is to meta-learn the most representative patterns from the whole video, adaptively and automatically.
These patterns are fed into ViT under a strict constraint of token length, therefore, they do not increase test time. Our method is based on bilevel optimization (i.e., meta-learning), and calls it as well as the resulted frames as \textbf{Meta-Optimized Frames (MOF)}. 
By incorporating MOF, video retrieval models are able to utilize the whole video for cross-modal inference, without increasing the number of input frames to ViT. As demonstrated in Figure~\ref{fig:example_exp}, MOF as a plug-in module improves the performance of the baseline method (Frozen \cite{bain2021frozen}) under different settings of input frames.

Technically for MOF, we perform two levels of optimization: \emph{model-level} (i.e., base-level in conventional meta-learning or bilevel optimization~\cite{mackay2019self,sinha2018review}) and \emph{frame-level} (i.e., meta-level in conventional meta-learning or bilevel optimization~\cite{mackay2019self,sinha2018review}). In our case, \emph{model-level} optimization takes the result of frame-level optimization (i.e., MOF) as input to ViT and aims to train the network parameters of ViT. \emph{Frame-level} optimization takes the input of ViT as ``parameters'', i.e., a selected number of video frames are parameterized to be learnable with gradient descent, and optimizes them (i.e., MOF) using the meta loss of ViT computed on the whole video. We formulate these two levels in a bilevel optimization program (BOP) \cite{mackay2019self,sinha2018review} that alternates the learning of two levels of optimization, through entire training phase. Each alternation is called a BOP phase.
In each phase, we perform a local BOP to distill the knowledge of regular frames into MOF. 
In specific, we first train a temporary model taking MOF as the input layer, where MOF in the beginning step is initialized with uniformly sampled frames. 
Then, we compute a validation loss with the whole video as input, and use the derived gradients to back-propagate and optimize the input layer (in training), i.e., the parameters of meta-optimized frames.
We conduct extensive experiments to evaluate the proposed MOF as a plug-in module in state-of-the-art methods, such as Frozen~\cite{bain2021frozen}, and CLIP4clip~\cite{luo2021clip4clip}.
Our superior results on three cross-modal retrieval benchmarks, e.g., $2.2\%$, $2.1\%$, and $1.9\%$ higher than Frozen~\cite{bain2021frozen} on R@10, demonstrate the effectiveness of MOF.

Our contributions are thus three-fold: (1) a novel meta-learning method called MOF to adaptively and automatically compress a video into fewer but more representative frames; (2) a novel BOP-based formulation and an end-to-end training solution that alternates between the learning of MOF and cross-modal video retrieval models; and (3) extensive experiments on three standard benchmarks to validate the effectiveness of MOF.

The rest of this paper is organized as follows. Section \ref{sec:rw} reviews related works. Section \ref{sec:pre} introduces the preliminaries of this video retrieval task. Section \ref{sec:meth} shows the details of our proposed method MOF. Section \ref{sec:exp} presents the extensive experiments and discussions on the results. Finally, Section \ref{sec:con} summarizes a few conclusions.

\section{Related work}\label{sec:rw}

In this section, we first briefly introduce the the progress of cross-modal video retrieval. Then, we review the most related
works from two aspects, i.e., one with a bilevel optimization program and the other with representative frame detection.

\subsection{Cross-modal video retrieval}

The core problem of cross-modal video retrieval is to measure the similarity between different modal features. A typical solution is to construct a common embedding space to learn video-textual correlations. According to training manners, we can divide existing works into two groups: video feature-based non-end-to-end and raw video-based end-to-end methods. 

\noindent
\textbf{Video feature-based non-end-to-end methods.} \cite{dong2021dual,chen2020fine, wu2021hanet, wang2020learning, yang2020tree, song2021spatial, wang2022many} are mainly based on video features generated via off-the-shelf video feature extractors as vision embedding and fed into the joint embedding space along with text to measure similarity. For example, Dong et al. \cite{dong2021dual, dong2019dual} adopt three branches, i.e. mean pooling, bi-GRU, and CNN to encode sequential videos and texts and learn a hybrid common space for video-text retrieval. Some studies have also explored rich multimodal information (e.g., motion, audio, and speech) \cite {mithun2018learning, liu2019use, gabeur2020multi, wang2021t2vlad} or large-scale datasets (e.g., HowTo100M) pre-training \cite {miech2020end, rouditchenko2020avlnet, patrick2020support} to improve the performance of cross-modal retrieval. Chen et al. and Wu et al. \cite{chen2020fine, wu2021hanet} propose fine-grained alignment models that decompose text and video into multiple levels and align text with video at multiple levels for video-text matching.

\noindent
\textbf{Raw video-based end-to-end methods.} \cite{lei2021less,bain2021frozen, wang2022object} train the model with raw video and paired text in an end-to-end manner. For instance, Lei et al. \cite{lei2021less} propose a general framework ClipBERT that enables end-to-end pre-training through a sparse sampling strategy. Bain et al. \cite{bain2021frozen} adopt a transformer-based video backbone and design a curriculum learning schedule to train the model on both image and video datasets to further improve performance. Inspired by Frozen \cite{bain2021frozen}, Wang et. al. \cite{wang2022object} present a transformer-based object-aware dual encoder model for end-to-end video-language pre-training. However, they both face the potential problem of reducing the high computational overload of dense video inputs.

Recently, pre-trained CLIP-based methods \cite {luo2021clip4clip,fang2021clip2video,cheng2021improving, zhao2022centerclip} have achieved superior results, thanks to the direct transfer of powerful knowledge from the pre-trained CLIP \cite{radford2021learning} and continued pre-training on a largescale video-language dataset. Different from these works, our work aims to achieve a decent trade-off between accuracy and speed by meta-learning efficient input frames. Our paradigm does not contradict pre-trained CLIP-based and transformer-based object-aware models \cite{luo2021clip4clip, wang2022object}. Instead, both pre-training CLIP-based and transformer-based object-aware models can be directly integrated into our training framework (e.g., as backbones) to gain extra improvements.

\subsection{Bilevel Optimization Program} 

Bilevel optimization program (BOP)~\cite{heinrich,goodfellow2014generative,wang2018dataset} aims to solve two levels of problems in one framework where an optimization problem contains another optimization problem as a constraint. A variety of problems arising in the area of machine learning can be formulated by bilevel optimization programs (BOP): incremental learning, adversarial training, and meta-learning. Recently, Liu et al. \cite{liu2021adaptive} propose a bilevel optimization-based approach for tackling the stability-plasticity dilemma in class incremental learning tasks. Zhu et al. \cite{zhu2019r2gan} redesign vanilla GAN \cite{goodfellow2014generative} can be formulated as a BOP,  which maximizes the reality score of generated images and minimizes the real-fake reconstruct loss.  Wei et al. \cite{wei2021meta} present a meta self-paced network that automatically learns a weighting scheme from data for cross-modal matching. In our work, we design a new version of BOP that alternatively optimizes the parameters of the retrieval models and the MOF across entire training phase. In each iteration, we apply a local BOP to learn the MOF specific to the input whole video.

\subsection{Representative Frame Detection}

Detecting representative frames from video input has been validated to be useful in video recognition and retrieval tasks \cite{mithun2017cmu,qiu2021condensing}. To be specific, Mithun et al. \cite{mithun2017cmu} used the key frames extracted by a dissimilarity-based sparse subset selection approach and evaluated it in the video-text retrieval task. Our work is also focused on learning representative frames for video retrieval tasks, and our key difference with \cite{mithun2017cmu} is: \cite{mithun2017cmu} extract keyframes only for video using a dissimilarity-based sparse subset selection approach  \cite{elhamifar2015dissimilarity}, while ours is a gradient descent based method in a bilevel optimization program which is built on the top of the base model (cross-modal model) to naturally incorporates text inputs. Bilen et al.  \cite{bilen2016dynamic} proposed to generate a dynamic image for each video by a rank pooling technique that captures the temporal evolution of actions. Its difference from ours is that it transforms the video to a single image with a manually designed method but it is not learnable (ours is learnable and thus optimal).

More recently, Tavakolian et al. \cite{tavakolian2019avd} also proposed to generate a single image from the video, and their method is based on adversarial learning. Qiu et al. \cite{qiu2021condensing} proposed to learn the transformation from the video to an informative frame and applied it to the task of video recognition. Our work is different from these two works mainly in two folds: 1) our method is flexible to generate any number (not only one) of images with a simple and elegant gradient descent process; and 2) our method works in an end-to-end manner with the “participation” of text inputs in the optimization process, that learns better “frames” for cross-modal retrieval tasks, i.e., between images and texts.

\begin{figure*}[t]
   \center
   \includegraphics[width=1\textwidth]{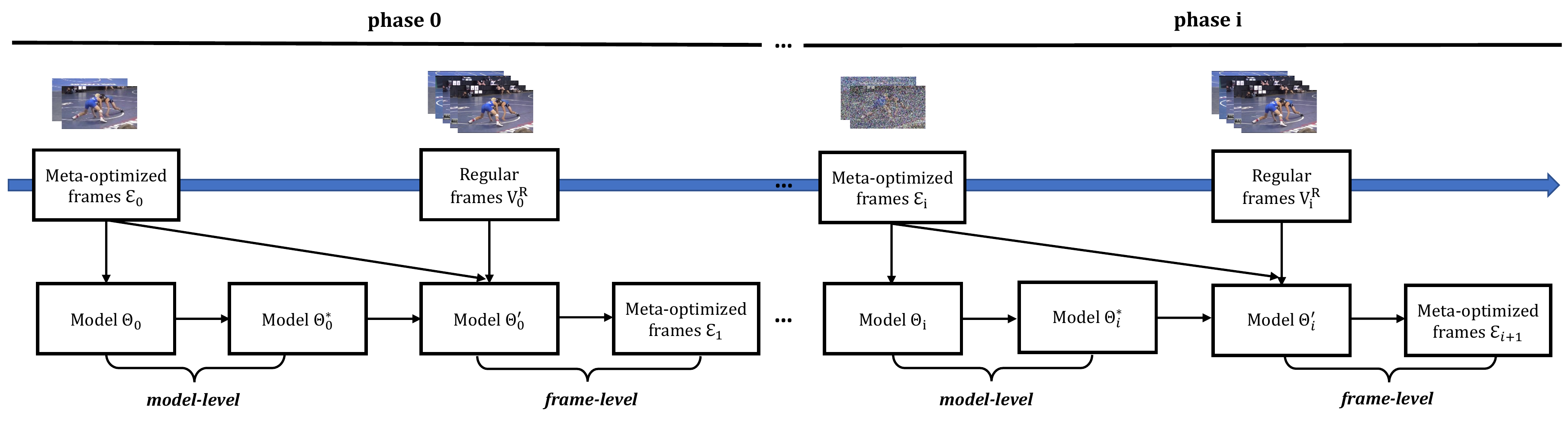}
   %\vspace{-0.1cm}
   \caption{The overall computing flow of the proposed MOF training. It is a global BOP that alternately learns the meta-optimized frames (we call frame-level optimization) and retrieval models (model-level optimization). The frame-level optimization within each phase is detailed in Figure 4.}
   \label{fig:framework}
   %\vspace{-0.3cm}
\end{figure*}

\section{Methodology}\label{sec:meth}

In this section, we first give the preliminary for cross-model retrieval. We then introduce the proposed Meta-Optimized Frames (MOF) in detail, including the global BOP, the model-level optimization, the frame-level optimization, and the algorithm flow.

\subsection{Preliminary}\label{sec:pre}

End-to-end video retrieval is proposed in \cite{bain2021frozen,lei2021less} to learn joint representations directly from video frame pixels and raw text tokens instead of offline-extracted single-modality features. We use $\rm \mathcal{P} = (V, T)$ to denote a video-text pair, where $\rm V \in \mathbb{R}^{N \times C \times W \times H}$ is a video, and $\rm T$ is its associated textual description, where $\rm C$ is color channels, $\rm (H, W)$ is the resolution of each raw frame. For a video clip $\rm V$, a typical approach is to uniformly sample a sequence of video frames $ \left \{ v_{1}, \cdot\cdot\cdot, v_{M}\right \}$ from $\rm V$ with a prespecified video temporal resolution, where $M$ is the length of the sampled video sequence. The prevailing cross-modal video retrieval model consists of a text encoder $F_{t}$ and a video encoder $F_{v}$. Given a text $\rm T$ and a video $\rm V$, $F_{t}$ and $F_{v}$ encode videos and textual descriptions as $E^{v}$ and $E^{t}$ respectively before mapping them to a joint embedding space, where the video-text similarity can be measured directly. With some similarity metrics, the encoded representations in $E^{v}$ and $E^{t}$ should be close if they are related, otherwise far apart.
% and far apart if not. 
Following \cite{bain2021frozen}, we employ a symmetrical contrastive loss \cite{zhai2018classification} in cross-modal video retrieval. 

The video-to-text loss $L_{v2t}$ and text-to-video loss $L_{t2v}$ then can be formulated as:
\begin{equation}\label{eqn:vtloss}
 \mathcal{L}_{v2t} = -\frac{1}{B} \sum^{B}_{i} \log \frac{\exp({E^v_i}^{\top} E^t_i/\sigma)}{\sum_{j=1}^{B} \exp({E^v_i}^{\top} E^t_j/\sigma)},
\end{equation} 
\begin{equation}\label{eqn:tvloss}
 \mathcal{L}_{t2v} = -\frac{1}{B} \sum^B_i \log \frac{\exp({E^t_i}^{\top} E^v_i /\sigma)}{\sum_{j=1}^B \exp({E^t_i}^{\top}E^v_j/\sigma)},
\end{equation} 
where $E^v_i$ and $E^t_j$ are the normalized embeddings of the $i$-th video and the $j$-th text respectively in a batch of size B, and $\sigma$ is the temperature. The overall loss function for training the retrieval model is the sum of the video-to-text loss ($\mathcal{L}_{v2t}$) and text-to-video loss ($\mathcal{L}_{t2v}$) :
\begin{equation}\label{eqn:loss}
\mathcal{L}_{c} = \mathcal{L}_{v2t} + \mathcal{L}_{t2v}. 
\end{equation} 
Recent breakthroughs in end-to-end training strategy with a single or a few sparsely sampled frames (or clips) indicate that the appropriate sparsely sampled strategy is the key to facilitating good and fast learning. In this paper, we aim to tackle the problem of cross-modal video retrieval by automatically compressing a video into fewer but more representative frames. Specifically,  we formulate this alternative learning with a new design of Bilevel Optimization Program (BOP) composed of both model-level and frame-level optimizations, specially for tackling cross-modal video retrieval tasks. After that, the details of the video and text encoder are further elaborated.

\textbf{Video and Text Encoder.} Early works usually use 2D/3D-CNN as a video encoder to extract offline video features \cite{dong2021dual,chen2020fine,song2021spatial}. Recently, several works \cite{bain2021frozen, zhang2021token, bertasius2021space} employ Visual Transformer(ViT) \cite{dosovitskiy2020image} to extract video feature in an end-to-end manner, which has shown promising performance in video modeling. The structure of ViT is similar to transformer in NLP tasks, which splits an input image into a sequence of patches and then flattens them into vectors (tokens) as inputs to transformer layers. In this work, we choose the dual-encoder framework Frozen \cite{bain2021frozen} as our base model to extract both video and text features in a trainable way. For the video encoder, we adopt Vision Transformer with space-time attention from TimeSformer \cite{bain2021frozen} as the backbone. More precisely, for the Vision Transformer, we use a ViT-B/16 model \cite{dosovitskiy2020image} with 12-layers and 768 widths with 12 attention heads. Then the output of [CLS] token at the final layer is extracted as the visual representation. For the text encoder, similar to previous methods \cite{bain2021frozen,lei2021less}, we choose transformer-based structure DistillBERT \cite{sanh2019distilbert} and treat the [CLS] token of the last hidden layer as the text representation. The video and text representation is later normalized by layer normalization and linearly projected into the joint video-text embedding space.

\section{Meta-Optimized Frames (MOF)}\label{sec:mof}

As illustrated in Figure \ref{fig:framework}, 
% the proposed MOF method alternates between the learning of a retrieval model and the learning of MOF in each learning phase. 
the proposed MOF method jointly works with a retrieval model in an alternative manner, in which one will update after the other's update phase.
The MOF are initialized with uniformly sampled frames of the input video, and then optimized while keeping the retrieval model temporarily fixed. Then, the MOF is fed into the retrieval model as input to train the model parameters. We formulate this alternative learning with a global BOP (Section \ref{sec:gb}), and name these two steps as model-level and frame-level optimizations, respectively. We provide the details of the proposed solutions, i.e., model-level and frame-level optimizations, in Sections \ref{sec:mlo} and \ref{sec:flo}, respectively. The training steps are elaborated in Section \ref{sec:alg}.

\subsection{Global BOP} \label{sec:gb} 

% The global BOP is inspired by the successful application of machine learning \cite{mackay2019self, liu2021adaptive}. The common retrieval model is optimized in each training epoch on regular frames; we propose this alternative learning strategy with a global BOP to use the (temporarily) optimal retrieval model to optimize MOF in each phase (i.e., model-level and frame-level training), and vice versa.

We can separate the global BOP into a sequence of bilevel optimization phases. In each phase, the retrieval model (denoted as $\rm {{\Theta}}$) is optimized on the MOF (denoted as ${\varepsilon}$), where 
% the MOF ${\varepsilon}$ is initialized by $\rm V^U$ ($\rm U \leq R$) and 
the parameters of MOF are meta-optimized by using a large set of regular frames (denoted as $\rm V^R$) sampled from the original video. 
%
% We use U and R to represent: the number of frames sampled for initialization and optimization of the MOF, respectively. 
Here, $\rm R$ denotes the number of sampled frames, and is much higher than the frame number of ${\varepsilon}$.
The MOF is trained based on the retrieval loss of $\rm V^R$.
It thus represents
the ``compressed'' information of $\rm V^R$.
Then, it is used as input to train the retrieval model.
% equivalent or more visual information with fewer frames. 
In this bilevel optimization process, the optimality of the retrieval model 
% derives 
endows
a constraint in optimizing MOF, and vice versa.

Specifically, in the $i$-th BOP phase,
% our retrieval system aims to
% learn a model 
we aim to learn $\rm {{\Theta}_i}$ to approximate the ideal model $\rm {{\Theta}^{\ast}_i}$, by minimizing a video retrieval loss $\mathcal{L}_c$ on the current MOF ${\varepsilon}_i$. We formulate this step as follows, 
\begin{equation}\label{eqn:theat}
\rm \Theta^{\ast}_i = \mathop{\arg \min} \limits_{\Theta_{i}} \mathcal{L}_{c}(\Theta_{i}(\varepsilon_{i}); V^R_i),
\end{equation} 
where $\varepsilon_{i}$ and $\rm V^R_i$ represent the meta-optimized frames (i.e., MOF) and the sampled regular frames in the $i$-th phase, respectively.

In the global BOP, the MOF $\varepsilon_{i}$ is optimized with the corresponding regular frames $\rm V^R_i$  through all phases. To learn more representative frames $\varepsilon^{\ast}_{i}$ from regular frames $\rm V^R_i$, our learning target for MOF is to minimize the model loss on these frames, i.e., $\rm V^R_i$. In particular, we define the objective of the global BOP as
\begin{equation}
\rm \mathop{\min} \limits_{\Theta_{i}} \mathcal{L}_{c}(\Theta_{i}( \varepsilon^{\ast}_{i}); V^R_i)
\end{equation} 
\begin{equation}
s.t.\ \ \rm \varepsilon^{\ast}_{i} = \mathop{\arg \min} \limits_{\varepsilon_{i-1}} \mathcal{L}_{c}(\Theta^{\ast}_{i-1}(\varepsilon_{i-1});  V^R_{i-1}),
\end{equation} 
where $\rm {\Theta^{\ast}_{i-1}}(\varepsilon_{i-1})$ denotes that the model $\rm{\Theta^{\ast}_{i-1}}$ was trained on $\varepsilon_{i-1}$ to transfer the knowledge from regular frames $\rm {V^R_{i-1}}$ into MOF $\varepsilon^{\ast}_i$, in the $(i-1)$-th phase. Please note that $\varepsilon^{\ast}_i$ will then be used as input in the $i$-th phase.

\subsection{Model-level Optimization}\label{sec:mlo} 

Figure~\ref{fig:framework} shows that in the $i$-th phase, we first solve the model-level optimization with 1) the MOF $\varepsilon_{i}$ as part of the input and 2) the model of the previous training phase $\rm {{\Theta}^{\ast}_{i-1}}$ as the initialization for model $\rm {{\Theta}_{i}}$. Let ${\alpha}$ be the learning rate of fine-tuning retrieval models, $\rm {{\Theta}_i}$ can be updated with one (or a few) gradient descent step(s) as
\begin{equation}
\rm {\Theta}^{\ast}_{i} \leftarrow {\Theta}_i - {\alpha} {\nabla}_{\Theta}\mathcal{L}_{c}(\Theta_i (\varepsilon_i)).
\end{equation} 
Then, $\rm {{\Theta}^{\ast}_i}$ will be temporarily fixed and deployed to learn MOF $\varepsilon_{i+1}$, i.e., to solve the frame-level optimization in the $i$-th phase, as elaborated in Section \ref{sec:flo}.

\subsection{Frame-level Optimization}\label{sec:flo}

In general, the number of frames fed into the retrieval model is considerably smaller than the total frame number of the original video $\rm V$. The existing methods \cite{bain2021frozen,lei2021less} made the assumption that models trained on a few frames achieve acceptable performance. However, this assumption cannot be guaranteed, especially when these frames are randomly or uniformly sampled. In other words, these sampled frames may not be representative or discriminative. In contrast, our approach based on optimization explicitly learns a feasible approximation to this assumption, given the differentiability of our MOF algorithm, and guarantees the representativeness of the ``sampled'' frames.

To achieve this, we initialize a temporary model $\rm{{\Theta}^{\prime}_i}$ (in the $i$-th phase) from model $\rm{{\Theta}^{\ast}_i}$ (i.e., $\rm{{\Theta}^{\prime}_i}=\rm{{\Theta}^{\ast}_i}$) and then train it on $\rm{{\varepsilon}_i}$ to maximize the prediction on regular frames $\rm {V^R_i}$. We use $\rm {V^R_i}$ to compute a validation loss to penalize this temporary training with respect to the parameters of $\rm{{\varepsilon}_i}$. The entire optimization is executed within a single stage, called local BOP, and can be formulated as
\begin{equation}
\rm \mathop{\min} \limits_{\varepsilon_{i}} \mathcal{L}_{c}({\Theta}^{\prime}_{i}( \varepsilon_{i}); V^R_i)
\end{equation} 
\begin{equation}
s.t.\ \  \rm {\Theta}^{\prime}_{i}( \varepsilon_{i}) = \mathop{\arg \min} \limits_{\Theta_{i}} \mathcal{L}_{c}(\Theta_{i};  \varepsilon_{i}).
\end{equation}

\begin{figure}[t]
   \center
   \includegraphics[width=0.46\textwidth]{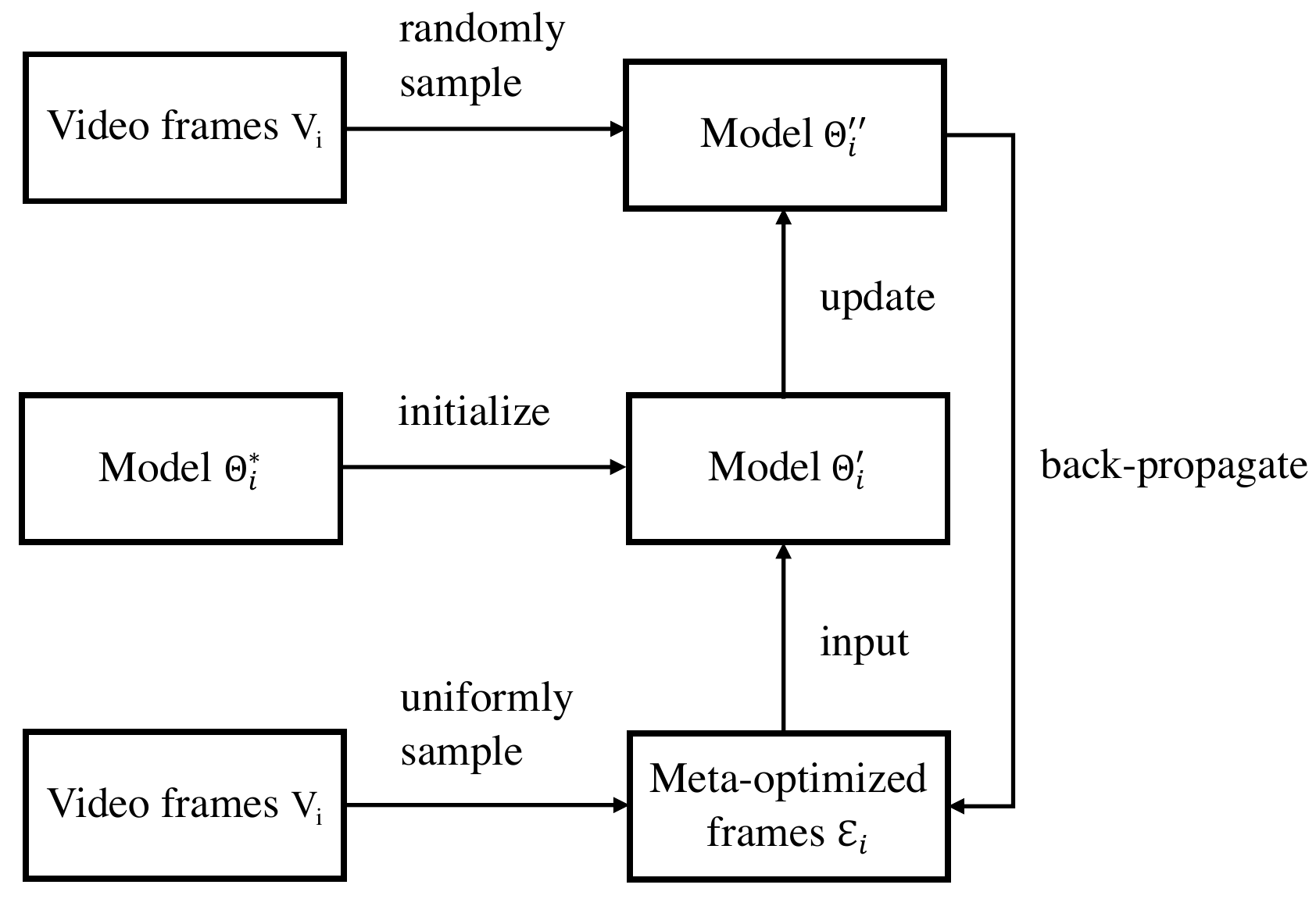}
   %\vspace{-0.1cm}
   \caption{uniform computing flow}
   \label{fig:bop}
   %\vspace{-0.2cm}
\end{figure}

\noindent
\textbf{Training $\rm{{\varepsilon}_i}$.} The training flow is detailed in Figure \ref{fig:bop}. First, 
the parameter size of
% the size of parameters 
$\rm{{\varepsilon}_i}$ (i.e., MOF) is initialized by $\rm {V_i}$. Second, we initialize a temporary model $\rm{{\Theta}^{\prime}_i}$ with $\rm{{\Theta}^{\ast}_i}$, and train $\rm{{\Theta}^{\prime}_i}$ on $\rm{\varepsilon_{i}}$ for one (or a few) iteration(s) by gradient descent as
\begin{equation}
\rm {\Theta}^{\prime \prime}_{i} \leftarrow {\Theta}^{\prime}_{i} - {\alpha} {\nabla}_{\Theta^{\prime}}\mathcal{L}_{c}(\Theta^{\prime}_{i};\varepsilon_{i}),
\end{equation}
where ${\alpha}$ is the learning rate of fine-tuning temporary models. Finally, as both $\rm{{\Theta}^{\prime}_i}$ and $\rm{{\varepsilon}_i}$ are differentiable, we are able to compute the loss of $\rm{{\Theta}^{\prime}_i}$ on $\rm {V^R_i}$, and back-propagate this validation loss to optimize $\rm{{\varepsilon}_i}$ by
\begin{equation}
\rm {{\varepsilon}_{i+1}} \leftarrow {\varepsilon}_i - {\beta} {\nabla}_{\varepsilon}\mathcal{L}_{c}(\Theta^{\prime \prime }_{i}({\varepsilon}_i); V^R_i),
\end{equation}
where ${\beta}$ is the learning rate. In this step, we
need to back-propagate the validation gradients till the input
layer, through unrolling all training gradients of $\rm{{\Theta}^{\prime}_i}$. This operation involves a gradient through a gradient. Computationally, it requires an additional backward pass through $\rm {\mathcal{L}_{c}(\Theta^{\prime}_{i};{\varepsilon}_i)}$ to compute Hessian-vector products, which is feasible by using deep learning toolboxes.

\subsection{Algorithm} \label{sec:alg}

In Algorithm 1, we summarize the entire process of the proposed MOF training. We show the alternative training between the retrieval models and the parameters of MOF, corresponding to Sections \ref{sec:gb} - \ref{sec:flo}. Specifically, for each phase, Steps 2-9 are for model-level training and Steps 12-16 are for frame-level training. Step 10 evaluates the learned model $\rm {\Theta_i}$ for the current phase, and the best value of all phases will be reported as the final evaluation.

\begin{algorithm}
\caption{Meta-Optimized Frames (MOF) Training}\label{alg}
\renewcommand{\algorithmicrequire}{\textbf{Input:}}
\renewcommand{\algorithmicensure}{\textbf{Output:}}
\begin{algorithmic}[1]
\Require  An untrimmed video $\rm V$, R: the number of regular frames, t: max phases, $\alpha$ and $\beta$: learning rates.
\Ensure Retrieval model $\rm {\Theta_i}$, and MOF ${\varepsilon}_i$.
\For{$i = 0, 1, 2,\cdot\cdot\cdot, t $}
    
    \State Get a real training video $\rm V_i$;

    \If{i=0}
        \State Uniform sampled frames from $\rm V_0$ to initialize $\varepsilon_0$;
        \State Initialize $\rm {\Theta_0}$ and train it on $\varepsilon_0$;
    \Else
        \State Get $\varepsilon_i$ from memory;
        \State Initialize $\rm {{\Theta}_i}$ with $\rm {{\Theta}^{\ast}_{i-1}}$;
        %\State Train $\rm {\Theta_i}$ on $\varepsilon_i$ by Eq. (7);
        \State Train $\rm {\Theta^{\ast}_i} \leftarrow {\Theta}_i - {\alpha} {\nabla}_{\Theta}\mathcal{L}_{c}({\Theta}_i(\varepsilon_i))$;
        \State Run test and record the results;
    \EndIf
    \State Randomly sampled frames $\rm V^R_i$ from $\rm V_i$;
    \State Initialize $\rm {{\Theta}^{\prime}_i}$ with $\rm {{\Theta}^{\ast}_{i}}$;
    \State Train $\rm {\Theta}^{\prime \prime}_{i} \leftarrow {\Theta}^{\prime}_{i} - {\alpha} {\nabla}_{\Theta^{\prime}}\mathcal{L}_{c}(\Theta^{\prime}_{i};\varepsilon_{i})$;
    \State Train $\rm {{\varepsilon}_{i+1}} \leftarrow {\varepsilon}_i - {\beta} {\nabla}_{\varepsilon}\mathcal{L}_{c}(\Theta^{\prime \prime}_{i}({\varepsilon}_i); V^R_{i})$;
    \State Update $\rm {{\varepsilon}_{i+1}}$ in memory.    
\EndFor
\end{algorithmic}
\end{algorithm}
\vspace{-0.2cm}

\section{Experiments} \label{sec:exp}

In this section, we conduct extensive experiments to verify the effectiveness of MOF on three widely-used cross-modal video retrieval datasets including MSR-VTT \cite{xu2016msr}, MSVD \cite{chen2011collecting}, and DiDeMo \cite{anne2017localizing}.

\subsection{Datasets}

\textbf{MSR-VTT}~\cite{xu2016msr} contains $10,000$ unique Youtube video clips with $20$ different text captions. Following other works~\cite{gabeur2020multi}, we use 9k train+val videos for training and report results on the 1k-A test set~\cite{Yu_2018_ECCV}. 

\textbf{MSVD}~\cite{chen2011collecting} contains $1,970$ clips from YouTube. Each video clip has around $40$ descriptions in multiple languages. Following previous work~\cite{venugopalan2015sequence}, there are $1,200$ clips for training, $100$ clips for validation, and $670$ clips for testing. 

\textbf{DiDeMo}~\cite{anne2017localizing} contains $10,000$ videos annotated with $40,000$ sentences. Following~\cite{liu2019use,bain2021frozen,lei2021less}, we evaluate video-paragraph retrieval, where all sentence descriptions for a video are concatenated into a single query.

\subsection{Evaluation Metrics}

We adopt the widely used median rank (MedR) and recall rate at top K (R@K) for assessing retrieval accuracy. MedR calculates the median rank position among where true positives are returned. R@K measures the fraction of true positives being ranked at the top K returned results. The higher R@K and lower MedR indicate better performance.

\subsection{Implementation Details}

\subsubsection{The architectures of $\rm{\Theta}$.} Following Frozen~\cite{bain2021frozen}, we use Timesformer~\cite{bertasius2021space} and multilayer bidirectional transformer as visual encoder and text encoder, respectively. For initialization of the model weights, we initialize the entire model with Frozen weights trained on the Google Conceptual Headings (CC3M)~\cite{sharma2018conceptual} dataset and the WebVid-2M dataset~\cite{bain2021frozen}.

\subsubsection{The architectures of $\rm{\varepsilon}$.} It depends on the size of the video frame and the number of video frames we need. On three datasets, we adopt video frame size 224 with patch size 16. The number of frames is set in two ways. (1) The MOF is uniformly sampled U frames from each video. Therefore, the parameter size of the frames per video is equal to H $\times$ W $\times$ U, where H, W, and C represent the height, width, and channel, respectively. (2) regular frames are randomly sampled R frames for each video. Similarly, the parameter size of the frames per video is equal to H $\times$ W $\times$ R. Notably, we use MOF and regular frames for training,  while using only regular frames at test time for efficient retrieval.

\subsubsection{Hyperparameters and configuration.} We implement our proposed method using PyTorch\footnote{\url{http://www.pytorch.org}} on 8 Tesla A100 GPUs (40G). For model-level hyperparameters, the model $\rm{\Theta_i}$ is finetuned with AdamW optimizer. In each (i.e., i-th) phase, the learning rate $\alpha$ is initialized as 1e-5. In Eq. 1 and 2, The temperature hyperparameter $\sigma$ is set to 0.05, following~\cite{bain2021frozen}. For frame-level hyperparameters, an Adam optimizer is used to update MOF $\rm{\varepsilon_i}$, using a learning rate $\beta$ of 8e-4. Following Frozen~\cite{bain2021frozen}, we use batch sizes of 16 when finetuning on all downstream datasets, i.e., cross-modal video retrieval datasets.

\subsection{Baselines}
To justify the effectiveness of our MOF, we compare it with the following thirteen methods: 
\begin{itemize}
\item \textbf{JSFusion \cite{yu2018joint}.} This is a cross-modal fusion method, which proposes a Joint Sequence Fusion model to fuse the video and text representation into a 3D tensor, and a convolutional network is employed to directly measure the similarity score. 
\item \textbf{Multi.Cues \cite{mithun2018learning}.} This classic method is also designed for video retrieval tasks, which utilizes multi-modal features by a fusion strategy and a weighted triplet ranking loss to better learn the joint embedding. 
\item \textbf{ActBERT \cite{zhu2020actbert}.} This is a unified multi-modal transformer framework, which encodes global actions, local regional objects, and text sentences in a Transformer to improve language-and-visual alignment.
\item \textbf{CE \cite{liu2019use}.} This is a multimodal fusion method, which uses collaborative gating to fuse rich multimodal features to obtain a better video representation.
\item \textbf{MMT \cite{gabeur2020multi}.} This is also a multimodal fusion approach, which uses BERT for text representation and proposes a multi-modal transformer to jointly encode diverse modalities in videos for video representation.
\item \textbf{AVNet \cite{rouditchenko2020avlnet}.} This is a multimodal retrieval approach, which uses raw audio and other modalities to better align and perform video-text retrieval.
\item \textbf{FSE \cite{zhang2018cross}.} This is a cross-model alignment approach, which proposes hierarchical modeling of videos and paragraphs. 
\item \textbf{T2VLAD \cite{wang2021t2vlad}.} This is also a cross-model alignment approach that introduces a paradigm of global-local alignment based on NetVLAD \cite{arandjelovic2016netvlad} to perform video retrieval. 
\item \textbf{SUPPORT-SET \cite{patrick2020support}.} This is a contrastive learning approach, which designs a generative objective to improve the instance discrimination limitations of contrastive learning. 
\item \textbf{TACo \cite{yang2021taco}.} This is also a contrastive learning approach that introduces token-aware cascade contrastive learning to improve the video-text alignment.
\item \textbf{ClipBERT \cite{lei2021less}.} This is an end-to-end retrieval approach, which leverages a sparse sampling strategy to train the model in an end-to-end manner.
\item \textbf{CLIP4Clip-meanP \cite{luo2021clip4clip}.} This is also an end-to-end retrieval approach, which uses CLIP \cite{radford2021learning} to extract the frame features and the text features, and then uses the mean pooling to aggregate the feature of all frames for video representation.
\item \textbf{Frozen \cite{bain2021frozen}.} This is an end-to-end pre-training approach, which uses the recent ViT \cite{arandjelovic2016netvlad} as the visual encoder and designs a curriculum learning schedule to train the model on both image and video datasets.

\end{itemize}

\setlength{\tabcolsep}{2.3mm}{
\renewcommand\arraystretch{1}
\begin{table*}
\caption{Text-to-video retrieval comparison with state-of-the-art methods on MSR-VTT. E2E: Works trained on raw videos in an end-to-end manner. Vis Enc. Init.: Datasets that visual encoders' initial weights are trained on. “\emph{w/} ours” means using our MOF training as a plug-in module. “SF” indicates the number of sampled frames. ``TT'' denotes the test time. The left and right of $\Rightarrow$ denote the number of sampled frames before and after meta-optimized frames learning strategy respectively.}
  \centering
  \begin{tabular}{lcllcccccc}
    \toprule
    \textbf{Method} & \textbf{E2E} & \textbf{Vis Enc.Init.} & \textbf{Pretrained Data} & \textbf{SF} & \textbf{TT(s)}
    & \textbf{R@1} & \textbf{R@5}  & \textbf{R@10} & \textbf{MedR}   \\
    
    \hline
    JSFusion  \cite{yu2018joint}     & \Checkmark &  - & - & -& - & 10.2               & 31.2   & 43.2  & 13.0                                 \\
    ActBERT  \cite{zhu2020actbert}     & \Checkmark &  VisGenome & HowTo100M & -& - & 16.3               & 42.8   & 56.9  & 10.0                                 \\
    CE  \cite{liu2019use}     & \XSolidBrush & Numerous experts & - & - & - & 20.9               & 48.8   & 62.4  & 6.0                                 \\
    MMT  \cite{gabeur2020multi}     & \XSolidBrush & Numerous experts & HowTo100M & - & - & 26.6               & 57.1   & 69.6  & 4.0                                 \\
    T2VLAD  \cite{wang2021t2vlad}     & \XSolidBrush & Numerous experts & HowTo100M & - & - & 29.5               & 59.0   & 70.1  & 4.0                                 \\
    AVLnet  \cite{rouditchenko2020avlnet}    & \XSolidBrush & ImageNet, Kinetics  & HowTo100M &- & -& 27.1               &  55.6  & 66.6  & 4.0                                        \\
    SUPPORT-SET  \cite{patrick2020support}    & \XSolidBrush & IG65M, ImageNet  & HowTo100M &- & -& 30.1               &  58.5  & 69.3  & 3.0                                        \\
    TACo  \cite{yang2021taco}    & \XSolidBrush &ImageNet, Kinetics  & HowTo100M & -&- & 26.7          &  54.5 & 68.2  & 4.0                                        \\
    ClipBERT  \cite{lei2021less}    & \Checkmark & - &COCO,VisGenome &16 &-& 22.0               & 46.8                & 59.9  & 6.0                                          \\
    Frozen  \cite{bain2021frozen}    & \Checkmark &ImageNet &  CC3M,WebVid-2M  & 8 & - & 31.0               & 59.5  & 70.5  & 3.0                                     \\
    Frozen  \cite{bain2021frozen} [Our Imp.]    & \Checkmark &ImageNet &  CC3M,WebVid-2M  & 8 & 46 & 31.3               & 59.6  & 70.6  & 3.0                                     \\
    \textbf{Frozen \emph{w/} ours}    & \Checkmark & ImageNet  & CC3M,WebVid-2M & 32 $\Rightarrow$ 8 & 46 & \textbf{33.8} &   \textbf{62.1}           & \textbf{72.8}  & \textbf{3.0}                                     \\
    \hline 
    CLIP4Clip-meanP  \cite{luo2021clip4clip} [Our Imp.]& \Checkmark &CLIP Weight & - & 12 & 5mins & 43.0 & 70.5       & 80.6  & 2.0     \\
    CLIP4Clip-meanP  \cite{luo2021clip4clip} [Our Imp.]& \Checkmark &CLIP Weight & -  & 4 & 4mins & 39.3       & 67.5  & 78.1 & 2.0                                      \\
    CLIP4Clip-meanP \emph{w/} ours  & \Checkmark &CLIP Weight & - & 12 $\Rightarrow$ 4 & 4mins &  40.5 &   68.2   & 79.5  & 2.0           \\       
    \bottomrule
  \end{tabular}
  \vspace{2mm}
  \label{tab:table_msr}
  %\vspace{-0.2cm}
\end{table*}
}

\setlength{\tabcolsep}{0.6mm}{
\renewcommand\arraystretch{1}
\begin{table}
\caption{Text-to-video retrieval comparison with state-of-the-art methods on MSVD. “\emph{w/} ours” means using our MOF training as a plug-in module. “SF” indicates the number of sampled frames. “TT” denotes the test time. The left and right of $\Rightarrow$ denote the number of sampled frames before and after meta-optimized frames learning strategy respectively.}
  \centering
  \begin{tabular}{lcccccccccc}
    \toprule
    \textbf{Method} & \textbf{SF} & \textbf{TT(s)} & \textbf{R@1} & \textbf{R@5}  & \textbf{R@10} & \textbf{MedR}  \\
    \hline
    Multi.Cues \cite{mithun2018learning}  & -&- &  20.3  & 47.8              & 61.1    & 6.0                                 \\
    CE \cite{liu2019use}& - & &  19.8               & 49.0   & 63.8  & 6.0                                 \\
    SUPPORT-SET \cite{patrick2020support} & - & -&  28.4               &  60.0  & 72.9  & 4.0                                        \\
    Frozen  \cite{bain2021frozen}& 8  & - & 33.7               & 64.7  & 76.3  & 3.0                                     \\
    Frozen  \cite{bain2021frozen} [Our Imp.] & 8  & 52 & 37.9               & 73.6  & 83.2  & 2.0                                     \\
    \textbf{Frozen \emph{w/} ours} & 32 $\Rightarrow$ 8 & 52 & \textbf{41.0}      & \textbf{74.1}   & \textbf{85.3}   &  \textbf{2.0}  \\ 
    % \midrule
    % \textbf{Method} & \textbf{SF} & \textbf{TT(s)} & \textbf{R@1} & \textbf{R@5}  & \textbf{R@10} & \textbf{TestT}  \\
    \hline
    CLIP4Clip-meanP [Our Imp.] & 12 & 3mins & 45.9 & 75.8  & 84.7  &  2.0           \\
    
    CLIP4Clip-meanP [Our Imp.] & 4 & 2mins & 43.1 & 72.7  & 82.3  &  2.0           \\
    
    CLIP4Clip-meanP \emph{w/} ours & 
    12 $\Rightarrow$ 4 & 2mins & 44.3 & 73.5  & 84.1  & 2.0                                      \\
    \bottomrule
  \end{tabular}
  %\vspace{2mm}
  \label{tab:table_msvd}
  \vspace{-0.4cm}
\end{table}
}

\setlength{\tabcolsep}{0.6mm}{
\renewcommand\arraystretch{1}
\begin{table}
\caption{Text-to-video retrieval comparison with state-of-the-art methods on DiDeMo. “\emph{w/} ours” means using our MOF training as a plug-in module. “SF” indicates the number of sampled frames. “TT” denotes the test time. The left and right of $\Rightarrow$ denote number of sampled frames before and after the meta-optimized frames learning strategy respectively.}
  \centering
  \begin{tabular}{lccccccccc}
    \toprule
    \textbf{Method} & \textbf{SF} & \textbf{TT(s)}  & \textbf{R@1} & \textbf{R@5}  & \textbf{R@10} & \textbf{MedR}  \\
    \hline
    FSE \cite{zhang2018cross}  &  - & - & 13.9  & 36.0   & -  & 11.0     \\
    CE \cite{liu2019use} & -  & - & 16.1               & 41.1   & -  & 8.3                                \\
    ClipBERT \cite{lei2021less} & 16  & - & 20.4              & 44.5               & 56.7  & 7.0                                          \\
    Frozen \cite{bain2021frozen} & 8   & - & 31.0  & 59.8  & 72.4  & 3.0                                     \\
    Frozen \cite{bain2021frozen} [Our Imp.] & 8  & 130  & 31.2  & 59.9  & 72.4  & 3.0                                     \\
    \textbf{Frozen \emph{w/} ours}  & 32 $\Rightarrow$ 8 & 130  & \textbf{33.6}             & \textbf{62.5}   & \textbf{74.3}  & \textbf{3.0}               \\
    
    \hline
    CLIP4Clip-meanP [Our Imp.] & 12 & 6mins & 43.5 &70.2   & 80.8   &   2.0         \\
    CLIP4Clip-meanP [Our Imp.] & 4 & 5mins & 40.5 &  67.3 &  78.2 &  2.0          \\
    
    CLIP4Clip \emph{w/} ours &  12 $\Rightarrow$ 4 & 5mins & 41.3 & 68.5  & 79.4  & 2.0                                      \\
    \bottomrule
  \end{tabular}
  %\vspace{2mm}
  \label{tab:table_didemo}
  \vspace{-0.4cm}
\end{table}
}

\setlength{\tabcolsep}{3.5mm}{
\renewcommand\arraystretch{1}
\begin{table*}
\caption{Text-to-video retrieval comparison with state-of-the-art methods on three datasets. “\emph{w/} ours” means using our MOF training as a plug-in module. “SF” indicates the number of sampled frames. The left and right of $\Rightarrow$ denote the number of sampled frames before and after meta-optimized frames learning strategy respectively.}
  \centering
  \begin{tabular}{lcccccccccc}
    \toprule
    \textbf{Dataset} & \textbf{Method} & \textbf{SF} & \textbf{R@1} & \textbf{R@5}  & \textbf{R@10} & \textbf{MedR} & \textbf{Train time(hrs)} & \textbf{Test time(s)}  \\
    \cmidrule(r){1-9}
    \multirow{8}*{MSR-VTT} 
    ~ & \multirow{3}*{Frozen \cite{bain2021frozen}[Our Imp.]} & 1 & 20.7 & 43.7  & 55.6  & 7.0  & 6 &    11 \\ 
    ~& ~ & 8 & 31.3 & 59.6  & 70.6  & 3.0  & 8  & 46 \\ 
    ~& ~ & 32 & 35.0 & 63.1  & 73.0  & 2.0  & 20 & 124 \\ 
    \cmidrule(r){2-9}
    ~& \multirow{5}*{Frozen \emph{w/} ours} 
    & 16 $\Rightarrow$ 1 & 22.1  & 46.2 & 57.5  & 7.0  & 14 & 12 \\        
    ~ & ~ & 8 $\Rightarrow$ 4  &  30.2 & 58.5   & 70.2 & 3.0 & 9 & 24  \\       
    ~ & ~ & 16 $\Rightarrow$ 4 & 31.1 & 59.3   & 71.1   & 3.0 & 17 & 24 \\      
    ~ & ~ & 8 $\Rightarrow$ 8 &  33.2 & 60.5  &  71.7 & 3.0  & 10 & 46 \\       
    ~ & ~ & 32 $\Rightarrow$ 8 & 33.8 &  62.1  & 72.8  & 2.0  & 28 & 46 \\                                          
    \cmidrule(r){1-9}
    \multirow{8}*{MSVD} 
    ~ & \multirow{3}*{Frozen \cite{bain2021frozen} [Our Imp.]} & 1 & 36.3  & 67.8 & 78.2  & 2.0  & 2.5 & 42    \\  
    ~& ~ & 8 & 37.9 & 73.6  & 83.2  & 2.0  & 4 & 52 \\ 
    ~& ~& 32 & 41.7 & 77.2  & 85.8  & 2.0  & 6 & 65 \\  
    
    \cmidrule(r){2-9}
    ~& \multirow{5}*{Frozen \emph{w/} ours}
      
     & 16 $\Rightarrow$ 1 & 36.4 & 68.5   & 79.3   & 2.0  & 4 & 42 \\     
    ~& ~ & 8 $\Rightarrow$ 4  & 38.5 & 73.2  & 82.5  & 2.0  &4.5  & 50  \\   
    ~& ~ & 16 $\Rightarrow$ 4 & 39.3 & 73.7   & 82.9   & 2.0  & 6 & 50 \\     
    ~& ~ & 8 $\Rightarrow$ 8 &  40.5 & 73.8  &  84.2 & 2.0  & 5  &  52  \\    
    ~& ~ & 32 $\Rightarrow$ 8 & 41.0 &  74.1  & 85.3  & 2.0  & 9  &  52 \\                     
    \cmidrule(r){1-9}
    \multirow{8}*{DiDeMo} 
     ~ & \multirow{3}*{Frozen \cite{bain2021frozen}[Our Imp.]} 
     & 1 &  21.5      & 47.3  & 60.3  & 6.0  & 10  & 80 \\  
     ~& ~ & 8 & 31.2       & 59.9  & 72.4   & 3.0  & 15  & 130  \\
     ~& ~ & 32 &  35.0      & 64.2  & 74.8 & 3.0  & 38 & 155 \\  
    
    \cmidrule(r){2-9}
    ~& \multirow{5}*{Frozen \emph{w/} ours}
     & 16 $\Rightarrow$ 1 & 22.0   & 49.0   & 61.2   & 6.0  & 16 &  80     \\
    ~& ~ & 8 $\Rightarrow$ 4  &30.8  & 59.5    & 72.0   & 3.0   & 21 & 95     \\
    ~& ~ & 16 $\Rightarrow$ 4 & 31.5  & 60.2   &72.5    &3.0   & 30  & 95      \\
    ~& ~ & 8 $\Rightarrow$ 8 & 32.3 & 60.7  & 72.6   & 3.0   & 25  & 130       \\
    ~& ~ & 32 $\Rightarrow$ 8 & 33.6  & 62.5   & 74.3  & 3.0  & 52 &  130       \\                                    
    \bottomrule
  \end{tabular}
  \vspace{2mm}
  \label{tab:table_abl}
  \vspace{-0.3cm}
\end{table*}
}

\setlength{\tabcolsep}{1.5mm}{
\renewcommand\arraystretch{1}
\begin{table}
\caption{Text-to-video retrieval comparison with different video backbones on MSR-VTT. “\emph{w/} ours” means using our MOF training as a plug-in module. Vis Enc.Init.: Datasets that visual encoders’ initial weights are trained on. “SF” indicates the number of sampled frames. The left and right of $\Rightarrow$ denote number of sampled frames before and after the meta-optimized frames learning strategy respectively. Note that for fair comparisons, all models were not pretrained on WebVid-2M and only finetuned on MSR-VTT train set.}
  \centering
  \begin{tabular}{lccccc}
    \toprule
    \textbf{Video backbone} & \textbf{Vis Enc.Init.} & \textbf{SF}  & \textbf{R@1} & \textbf{R@5}  & \textbf{R@10}   \\
    \hline 
    ResNet-50 & ImageNet-1k & 2  &  2.0  & 5.0  & 10.4      \\
    
    ResNet-50 \emph{w/} ours & ImageNet-1k &  8 $\Rightarrow$ 2 &  2.4 & 7.3   & 13.6     \\
    
    ResNet-152 & ImageNet-1k  & 2  & 2.3 & 5.9  &  10.5                      \\
    ResNet-152 \emph{w/} ours & ImageNet-1k & 8 $\Rightarrow$ 2  &  3.3 & 10.4   & 15.0                        \\
    TimeSformer   & ImageNet-21k & 2 &  14.0              & 36.9              & 52.0                                         \\
    TimeSformer \emph{w/}  ours  & ImageNet-21k  & 8 $\Rightarrow$ 2 & 15.1             & 41.7  & 54.3               \\
    
    \bottomrule
  \end{tabular}
  %\vspace{2mm}
  \label{tab:table_abl2}
  \vspace{-0.4cm}
\end{table}
}

\subsection{Experimental Results and Analyses}

\subsubsection{Comparisons with State-of-the-art} We compare the results on text-to-video retrieval in Tables~\ref{tab:table_msr}, \ref{tab:table_msvd}, and \ref{tab:table_didemo}. The results show that taking our method as a plug-in module on the state-of-the-art~\cite{bain2021frozen} consistently yields better performance on three datasets. To ensure a fair comparison, we re-implement the recent method Frozen~\cite{bain2021frozen} with our downloaded MSVD and DiDeMo datasets. Notably, improvements can be as high as $2.2\%$, $2.1\%$, $1.9\%$ relative to the Frozen~\cite{bain2021frozen} on R@10 for MSRVTT, MSVD, and DiDeMo, respectively. To further compare our approach with recent CLIP-based work~\cite{luo2021clip4clip}, we take our MOF as a plug-in architecture for the CLIP4Clip method~\cite{luo2021clip4clip}. From Tables \ref{tab:table_msr}, \ref{tab:table_msvd}, and \ref{tab:table_didemo},  we can see that CLIP4Clip \emph{w/} ours achieves comparable results with fewer frames and shorter test times. Furthermore, we observe that raw video-based end-to-end methods~\cite{bain2021frozen,luo2021clip4clip} achieve better performance than video feature-based non-end-to-end methods \cite{gabeur2020multi,patrick2020support}, showing the advantage of pre-trained visual transformers on video representation.

\subsubsection{Ablation Study} We conduct extensive ablation study on the number of sampled frames and different video backbones as shown in Tables~\ref{tab:table_abl} and \ref{tab:table_abl2}.

\vspace{0.1cm}
\noindent \textbf{Number of sampled frames.}  In Table \ref{tab:table_abl}, we compare the performance of Frozen \emph{w/} ours by varying the number of frames on the MSR-VTT, MSVD, and DiDeMo datasets, respectively. In Table 4, we observe that the boost by our MOF training becomes larger in more-frame settings, e.g. on MSR-VTT, Frozen \emph{w/} ours gains $71.7\%$ in R@10 (SF= 8 $\Rightarrow$ 8) while $72.8\%$ in R@10 (SF= 32 $\Rightarrow$ 8). This condition reveals that our MOF training yields better performance by compressing more frames. For MSR-VTT, compared to fewer-frames settings, our Frozen \emph{w/} ours (SF= 8 $\Rightarrow$ 4) achieves comparable performance (70.2) with Frozen (70.6) under the same parameter settings but using fewer frames (4 \emph{vs} 8). Additionally, performances further increase when compressing more frames (70.2$\rightarrow$71.1). Another observation is that our Frozen \emph{w/} ours (SF= 8 $\Rightarrow$ 4) has slightly more training time (9 \emph{vs} 8) and less testing time (24 \emph{vs} 46) than Frozen (SF=8). Thus, our Frozen \emph{w/} ours can make a good trade-off between effectiveness (i.e., retrieval accuracy) and efficiency (i.e., computing speed), which have a significant impact on real-world problems such as cross-modal video search online.  Finally, we compare Frozen \emph{w/} ours (SF= 16 $\Rightarrow$ 1) with Frozen (SF=8) on MSR-VTT. We report that this variant of our model produced results about 13.1\% worse than Frozen. We conjecture that the performance decrease with 1 frame is due to the reduced temporal information. Due to GPU memory constraints, we are not able to train any models with R values higher than 32 as those models have a much higher computational cost. Analogously, our method achieves a new state-of-the-art performance on the MSVD and DiDeMo datasets.

\vspace{0.1cm}
\noindent \textbf{Different video backbones.} The core of our proposed MOF is a novel meta-learning-based training paradigm that can be adapted to different video backbones. Therefore, to explore the robustness of our proposed MOF regarding different video backbones, we conduct corresponding analysis experiments. In MSR-VTT, we use three types of backbones as the video backbone, including ResNet-50, ResNet-152, and TimeSformer. The ablation of the different video backbones are in Table \ref{tab:table_abl2}. For fair comparisons, all models were not pretrained on WebVid-2M and only finetuned on the MSR-VTT train set. From the results, we find that taking our MOF as a plug-in module on TimeSformer achieves the best performance. We also note that ResNet-50  and ResNet-151 also achieve better results when being used together with our MOF, which indicates that our MOF can be plugged into different video backbones.

\begin{figure*}[t]
    \centering
    \subfloat[]{\includegraphics[width=0.5\textwidth]{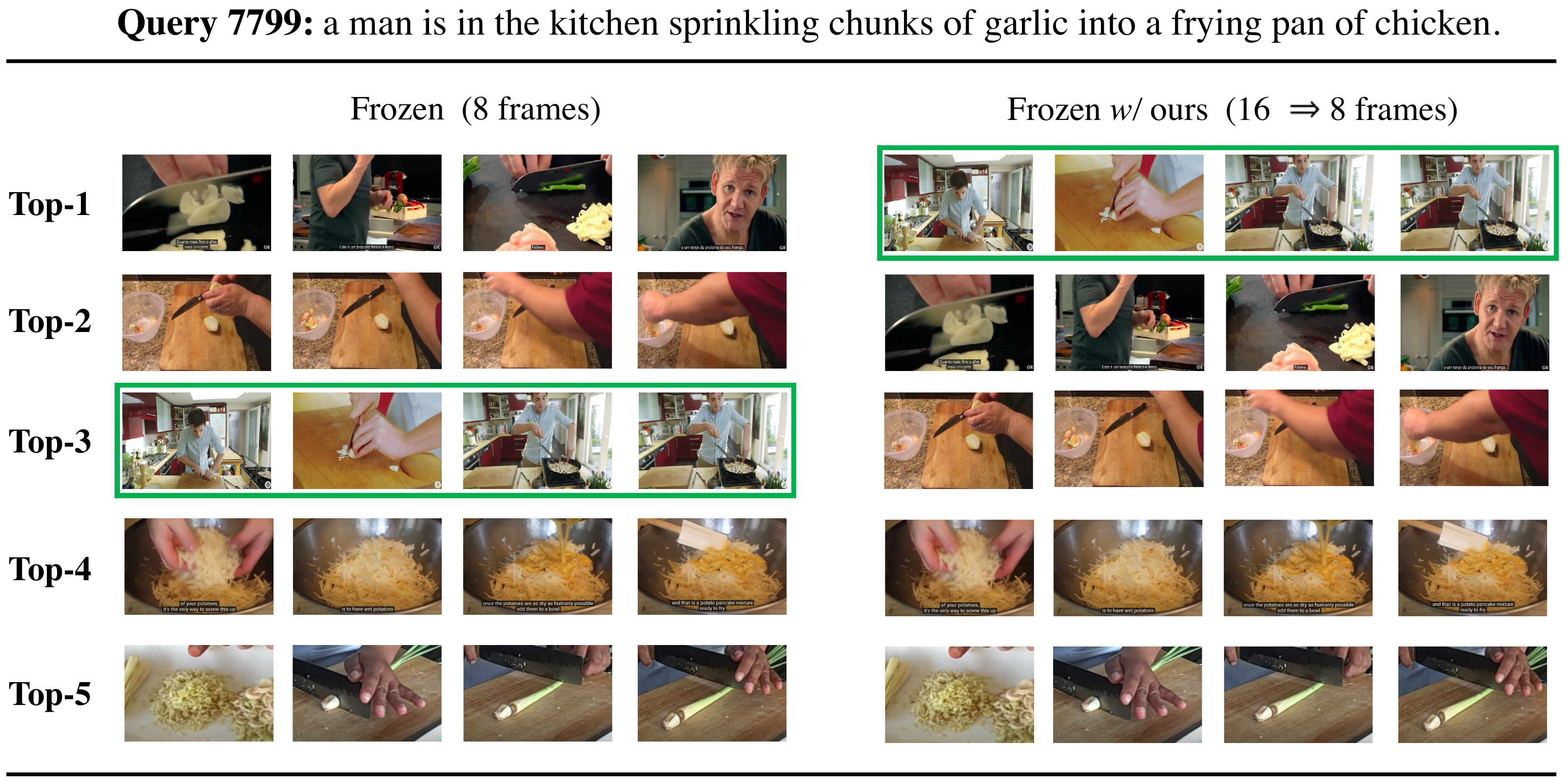}
    \label{fig_alpha1}}
    %\hfil
    \subfloat[]{\includegraphics[width=0.5\textwidth]{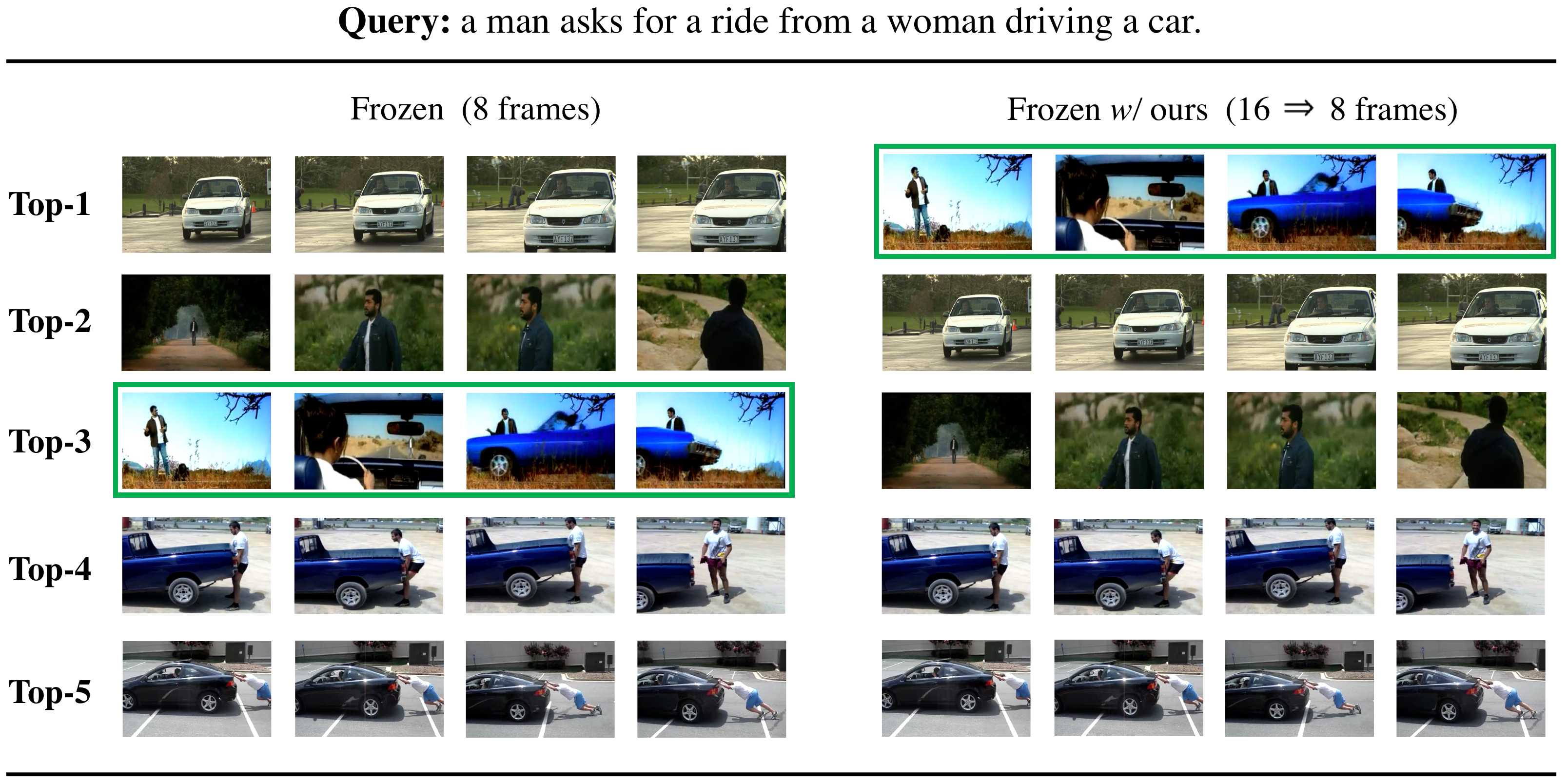}
    \label{fig_alpha2}}
    %\vspace{-0.3cm}
    \caption{Qualitative examples of text-to-video retrieval. In (left), (right), we show the Top-5 ranks of Frozen \cite{bain2021frozen} and  our Frozen \emph{w/} ours on the MSR-VTT and MSVD datasets. Given a textual description as a query, we retrieve the most relevant video ranked from top to bottom. Ground-truth video is bounded in green box.}
    \label{fig:qualitative_vis}
    \vspace{-0.2cm}
\end{figure*}

\begin{figure}[t]
   \center
   \includegraphics[width=0.5\textwidth]{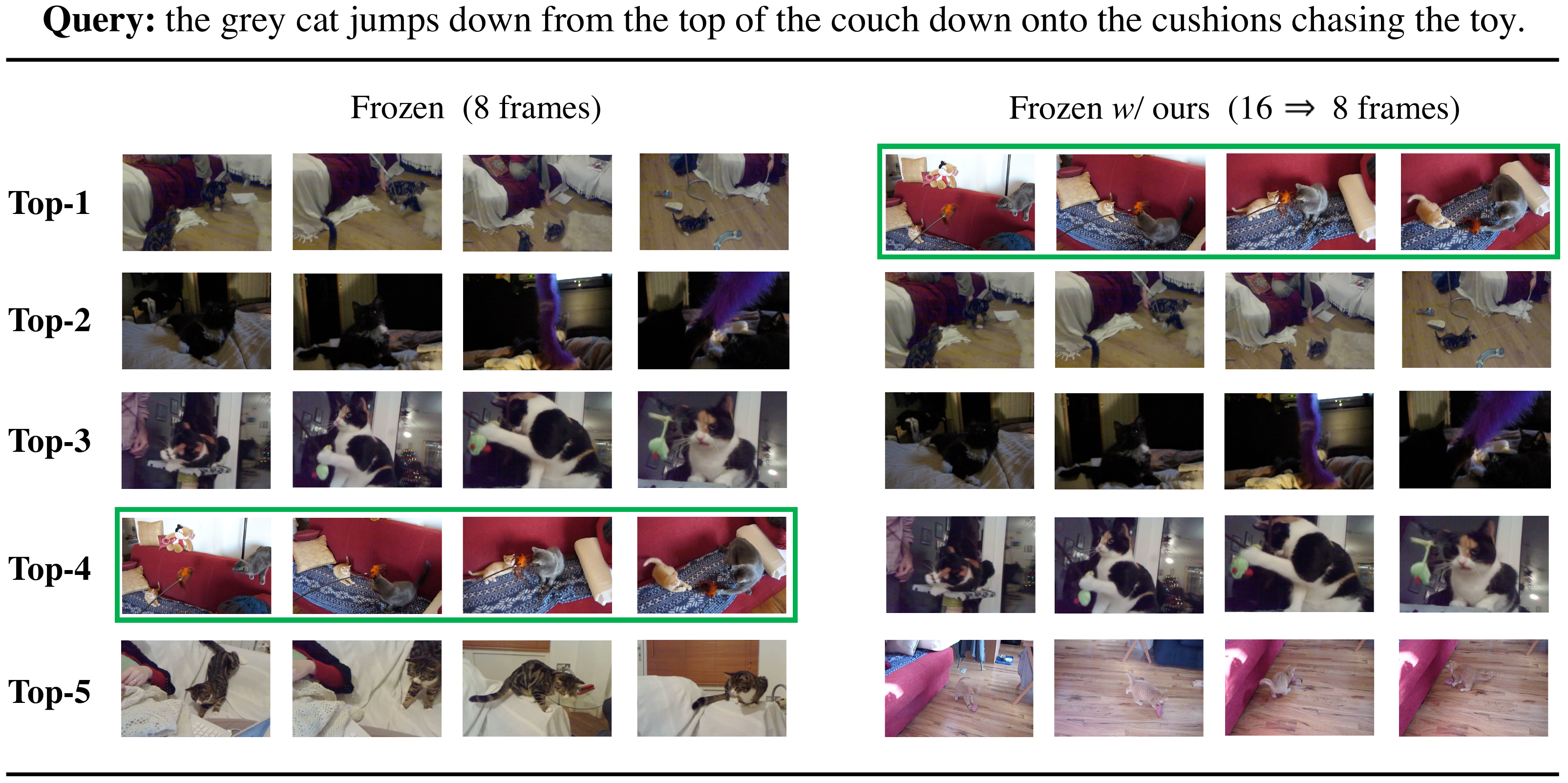}
   %\vspace{0.1cm}
   \caption{Qualitative examples of text-to-video retrieval. In (left), (right), we show the Top-5 ranks of Frozen \cite{bain2021frozen} and  our Frozen \emph{w/} ours on the DiDeMo dataset. Given a textual description as a query, we retrieve the most relevant video ranked from top to bottom. Ground-truth video is bounded in green box.}
   \label{fig:qualitative_didemo}
   %\vspace{-0.3cm}
\end{figure}

\subsubsection{Qualitative Analysis} In Figure~\ref{fig:qualitative_vis} (a), we visualize two qualitative examples of text-to-video retrieval results between Frozen and Frozen \emph{w/} ours on MSR-VTT. The text query describes multiple objects (e.g., “man”, “garlic”, and “pan”), and the action (“sprinkling”) in a short-term segment. In Figure 5 (right), the two videos ranked at the top contain a similar set of objects and actions. By using MOF, our Frozen \emph{w/} ours successfully ranks the ground-truth video at the top. Figure 5 (left) shows an example where Frozen with 8 frames cannot distinguish videos with similar objects where action attributes are predicted wrongly. In Figure 5 (left), the top-1 and top-2 of the retrieved videos both contain the “man” and “kitchen” objects and lack the “sprinkling chunks of garlic into a frying pan” action. The top-4 and top-5 of the retrieved videos do not contain the “garlic” object and its action attributes (e.g., sprinkling) are correctly predicted. The performance of Frozen \emph{w/} ours indicates that learning MOF enhances the expressiveness of the video representation and further improves the retrieval performance.

In Figures \ref{fig:qualitative_vis} (b) and \ref{fig:qualitative_didemo}, we provide additional qualitative text-to-video retrieval results on MSVD~\cite{chen2011collecting} and DiDeMo~\cite{anne2017localizing}. Given a text query, in most cases, our Frozen \emph{w/} ours successfully ranks the true positive in the top-1 position. In Figure \ref{fig:qualitative_vis} (b) (right), our Frozen \emph{w/} ours can retrieve top-1 video that contains relevant semantics (e.g., “man” and “car”) from the textual query. Figure \ref{fig:qualitative_vis} (b) (left) shows the retrieved top-1 and top-2 videos where some key elements (“man” or “car”) and their objects and motions are incorrectly predicted. This is mainly because sufficient visual information is missed in the sampled frames. On DiDeMo, Figure \ref{fig:qualitative_didemo} also shows that the results are in line with the conclusions of MSR-VTT and MSVD. Hence, the proposed MOF is effective in cross-modal video retrieval. 

\begin{figure*}[t]
   \center
   \includegraphics[width=0.9\textwidth]{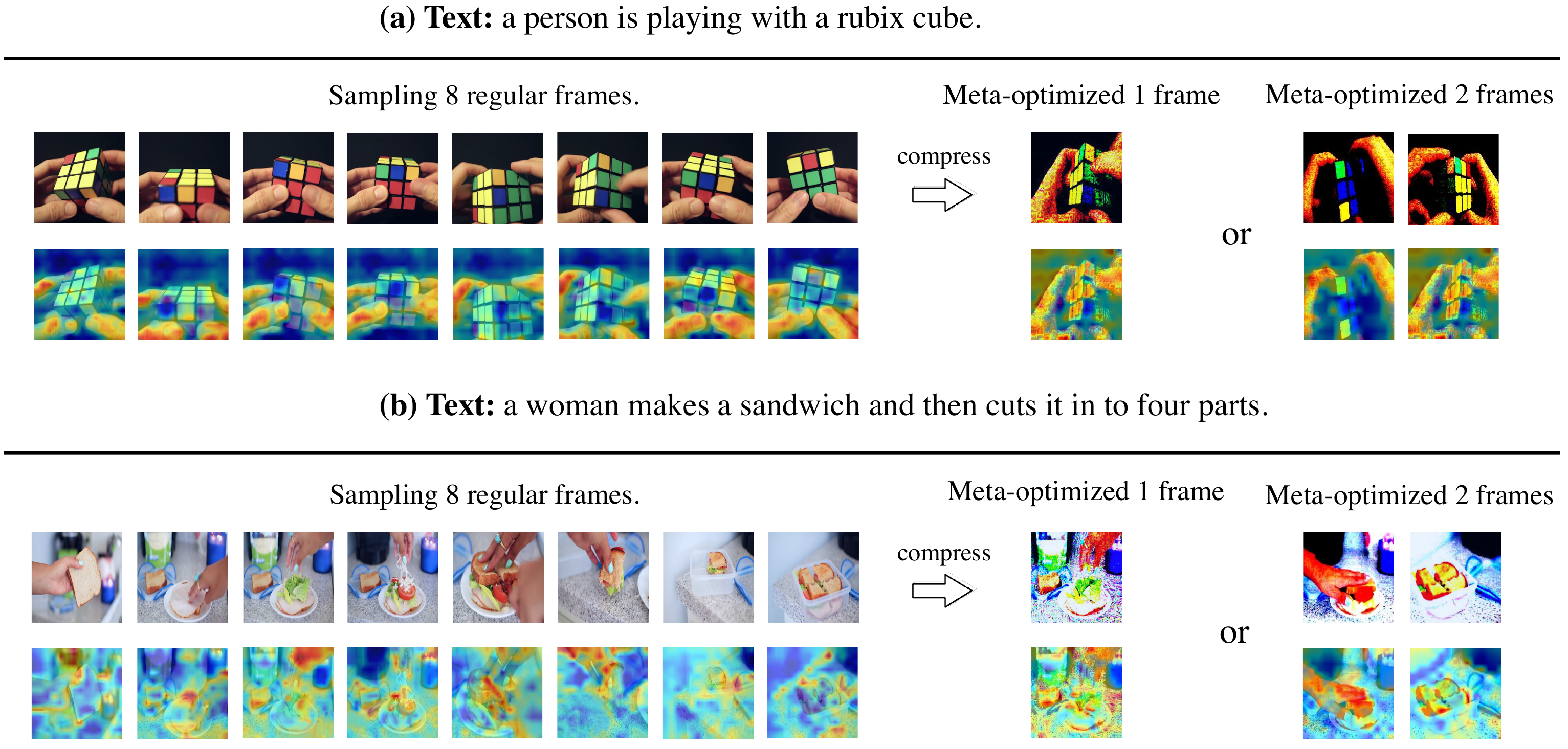}
   %\vspace{-0.1cm}
   \caption{Visualization of meta-optimized frames and cross-modal attention maps on sample frames from the MSR-VTT dataset. Samples are selected from the training set coming in the 0-th phase (left) and the 50-th phase (right), respectively. The original 8 frames are shown on the top left, and the corresponding meta-optimized 1 frame on the top right. We also show the corresponding attention maps separately at the bottom.}
   \label{fig:frames_vis}
   %\vspace{-0.3cm}
\end{figure*}

\begin{figure}[t]
   \center
   \includegraphics[width=0.5\textwidth]{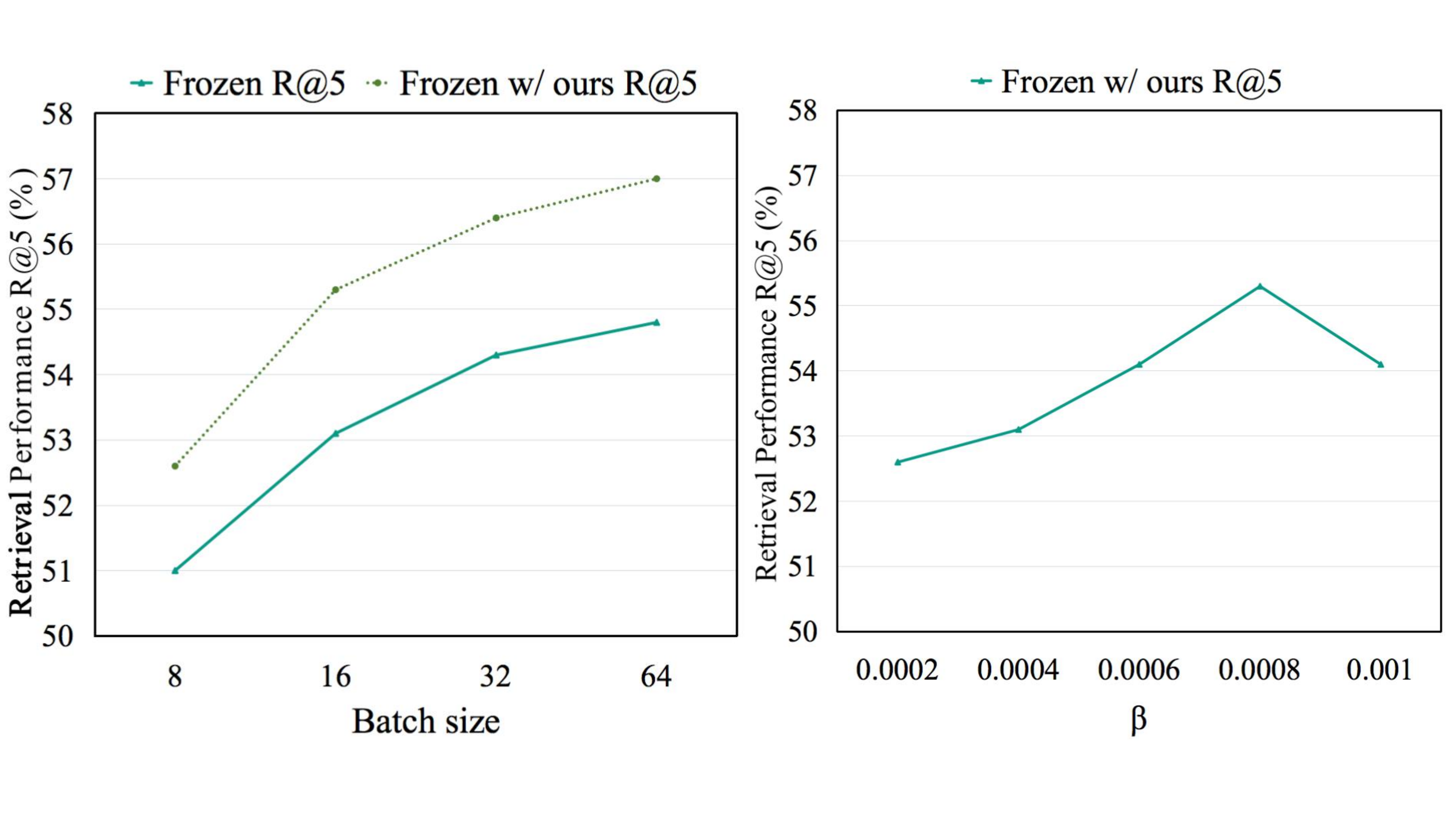}
   %\vspace{-0.1cm}
   \caption{Retrieval performance on different batch sizes and learning rates $\beta$.}
   \label{fig:batch_size}
   %\vspace{-0.2cm}
\end{figure}

\subsubsection{Visualization Results} To intuitively observe the effectiveness of introducing MOF training, we visualize the attention map between video frames and corresponding sentences. We visualize the attention map from the output of the first Transformer layer. Specifically, we use a text token as the query and visualize the attention weights on all spatial tokens. To analyze if MOF can enhance the expressiveness of the video representation, in Figures 6 (a) and (b), we visualize the regular frames and the MOF, where regular frames are from different scenarios. We can see that our Frozen \emph{w/} highlights key visual cues related to the text, e.g., human hand, sandwich, and cooking tools, and ignores (the useless) background pixels such as home decorations.  Furthermore, the MOF has significantly more salient regions (orange) than each regular frame. Visualization verifies that the learned MOF provides more useful visual semantic information, thus boosting the expressiveness of the video representation. This is the essence behind our video compression idea.

5) Parameter Sensitivity: As shown in Figure \ref{fig:batch_size}, we conduct parameter sensitivity experiments in our model. Specifically, we evaluate the impact of the parameters batch size and learning rate $\beta$ in Eq.11. Notably, we omit the text-video retrieval results on the other two datasets due to space limitations, which show similar trends to MSR-VTT~\cite{xu2016msr}. To analyze the influence of the hyperparameter, that is, batch size on the MSR-VTT datasets, we train Frozen \cite{bain2021frozen} with 2 frames and Frozen \emph{w/} ours (i.e., using our MOF training as a plug-in module) with 16 frames compressed into 2 frames. Figure \ref{fig:batch_size} (left) presents the results across batch sizes on the MSR-VTT dataset by Recall@5; note that Recall@1 and Recall@10 present the same trend. We observe that increasing the batch size leads to consistent performance gains. Due to GPU memory constraints, we are not able to test our model on batch sizes larger than 64. Note that the curves of each parameter are obtained by fixing the remaining other parameters. To explore the effect of the learning rate $\beta$, we train Frozen w/ ours with a batch size of 16 to compress 16 frames into 2 frames on the MSR-VTT dataset. Figure \ref{fig:batch_size} (right) demonstrates that the best performance for the retrieval performance is achieved when $\beta = 0.0008$. Moreover, we also observe that the proposed method is more sensitive to $\beta$, which demonstrates the importance of the learning rate $\beta$ for meta-optimized frames.

\section{Conclusions}\label{sec:con}

In this paper, we proposed a novel meta-learning method Meta-Optimized Frames (MOF) that automatically compresses the video into fewer but more representative frames.
% We call this method Meta-Optimized Frames (MOF) training framework for 
% tackling cross-modal video retrieval tasks. 
% Our main contribution is the method of 
In essence, our learned ``meta-optimized frames'' are
% not only efficient frames but also 
optimizable and adaptable, and can improve the training flexibility of online retrieval systems. 
We conducted extensive experiments on three video retrieval benchmarks and validated that our MOF achieves competitive multi-modal retrieval performance while greatly reducing the inference costs. 
In addition, our training method of MOF is generic and could be potentially incorporated into other video tasks, e.g., video captioning and video classification, to improve the model performance. We will take these as our future works.

\section*{Acknowledgments}
This research is supported by A*STAR under its AME YIRG grant (Project No. A20E6c0101).

% Can use something like this to put references on a page
% by themselves when using endfloat and the captionsoff option.
\ifCLASSOPTIONcaptionsoff
  \newpage
\fi

% trigger a \newpage just before the given reference
% number - used to balance the columns on the last page
% adjust value as needed - may need to be readjusted if
% the document is modified later
%\IEEEtriggeratref{8}
% The "triggered" command can be changed if desired:
%\IEEEtriggercmd{\enlargethispage{-5in}}

% references section

% can use a bibliography generated by BibTeX as a .bbl file
% BibTeX documentation can be easily obtained at:
% http://mirror.ctan.org/biblio/bibtex/contrib/doc/
% The IEEEtran BibTeX style support page is at:
% http://www.michaelshell.org/tex/ieeetran/bibtex/
\bibliographystyle{IEEEtran}
% argument is your BibTeX string definitions and bibliography database(s)

%
% <OR> manually copy in the resultant .bbl file
% set second argument of \begin to the number of references
% (used to reserve space for the reference number labels box)

\bibliography{IEEEfull,ref}

% biography section
% 
% If you have an EPS/PDF photo (graphicx package needed) extra braces are
% needed around the contents of the optional argument to biography to prevent
% the LaTeX parser from getting confused when it sees the complicated
% \includegraphics command within an optional argument. (You could create
% your own custom macro containing the \includegraphics command to make things
% simpler here.)
%\begin{IEEEbiography}[{\includegraphics[width=1in,height=1.25in,clip,keepaspectratio]{mshell}}]{Michael Shell}
% or if you just want to reserve a space for a photo:

% You can push biographies down or up by placing
% a \vfill before or after them. The appropriate
% use of \vfill depends on what kind of text is
% on the last page and whether or not the columns
% are being equalized.

\vfill

% Can be used to pull up biographies so that the bottom of the last one
% is flush with the other column.
%\enlargethispage{-5in}

%\appendix

%\input{MOF/Figs/Fig5_qualitative_vis_msvd}

%\input{MOF/Figs/Fig5_qualitative_vis_didemo}

% that's all folks
\end{document}